\documentclass{article}% single-column arXiv preprint style
\usepackage{arxiv}

\usepackage[utf8]{inputenc}
\usepackage[T1]{fontenc}
\usepackage[colorlinks=true,linkcolor=black,citecolor=black,urlcolor=blue,filecolor=black]{hyperref}
\usepackage{natbib}
\usepackage{graphicx}
\usepackage{amsmath}
\usepackage{amssymb}
\usepackage{booktabs}
\usepackage{array}

\newcommand{\xF}{$\times$F}          % frontier-normalized multiple
\usepackage{xcolor}
\usepackage{url}
\renewcommand{\headeright}{Preprint}
\renewcommand{\undertitle}{A Preprint}
\renewcommand{\shorttitle}{Closed-Loop Evaluation of ED Trajectory Simulators}

\makeatletter
\renewcommand{\@maketitle}{%
  \vbox{%
    \hsize\textwidth
    \linewidth\hsize
    \vskip 0.1in
    \@toptitlebar
    \centering
    {\LARGE\bfseries \@title\par}
    \@bottomtitlebar
    \textsc{\undertitle}\\
    \vskip 0.1in
    \def\And{%
      \end{tabular}\hfil\linebreak[0]\hfil%
      \begin{tabular}[t]{c}\bfseries\rule{\z@}{24\p@}\ignorespaces%
    }%
    \def\AND{%
      \end{tabular}\hfil\linebreak[4]\hfil%
      \begin{tabular}[t]{c}\bfseries\rule{\z@}{24\p@}\ignorespaces%
    }%
    \begin{tabular}[t]{c}\bfseries\rule{\z@}{24\p@}\@author\end{tabular}%
    \vskip 0.4in \@minus 0.1in
    \center{\@date}%
    \vskip 0.2in
  }%
}
\makeatother

\title{What Next-Event Accuracy Cannot See:\\
Closed-Loop Evaluation of Emergency Department\\
Trajectory Simulators}

\author{
  Zhen Xuen Brandon Low\\
  Monash University\\
  \texttt{zlow0014@student.monash.edu}
}

\date{}

\begin{document}

\maketitle

\begin{abstract}
Clinical trajectory models are usually evaluated by next-event accuracy on observed
histories. Simulation is different: models must condition on their own generated
events, allowing errors to compound. Although this problem is well known in sequence
modelling, it has not been systematically quantified for clinical trajectory
simulators. We developed EDSim-Bench to evaluate this failure mode using 425{,}028
MIMIC-IV-ED stays, with external replication on MC-MED, and release the evaluation
protocol and scoring code. Starting from held-out visit prefixes, models generate the
remainder of each visit and are evaluated on termination, event composition, timing,
conditional fidelity, and occupancy forecasting, with a train-only order-3 $n$-gram
as a reference baseline. Despite next-event accuracies within 0.001, three neural
architectures behaved very differently under rollout. Across seeds, one Transformer
recipe ranged from 0.43 to 0.96 in termination score and from 4.2- to 137-fold the
divergence of the $n$-gram; no prefix-trained neural model approached the $n$-gram on
termination or event composition. Inference-time interventions improved termination
but did not jointly recover composition and timing. Supervising every eligible
sequence position rather than only the final prefix position was associated with one
to two orders of magnitude lower divergence across Transformer, GRU, and LSTM models,
with the pattern persisting under model scaling, temporal shift, and external-site
evaluation. Nevertheless, even the best model generated visits approximately half as
long as observed, and model rankings reversed on occupancy forecasting, a downstream
quantity relevant to bed management. These results show that next-event accuracy is
insufficient to evaluate clinical trajectory simulators and motivate closed-loop
evaluation across seeds, rollout criteria, and downstream tasks.
\end{abstract}

\keywords{closed-loop evaluation, generative EHR models, trajectory simulation, exposure bias, emergency department, benchmark}

\paragraph*{Data and Code Availability}
MIMIC-IV-ED 2.2, MIMICEL 2.1.0, and MC-MED 1.0.1 are de-identified,
credentialed-access PhysioNet datasets \citep{johnson2023mimicived,wei2023mimicel,
kansal2025mcmed}. Code for deterministic cohort and split reconstruction, model
training, closed-loop rollout, and scoring is available at
\url{https://github.com/zxKyouma/EDSim-Bench}.
Source-derived prefixes, checkpoints, and count-model artifacts are not
redistributed; credentialed users can reconstruct them with the released code.

\paragraph*{Institutional Review Board (IRB)}
This secondary analysis used only de-identified datasets accessed under their data-use
agreements and did not require additional IRB review; the source datasets were released
under their respective institutional approvals.

\section{Introduction}
\label{sec:intro}

Clinical trajectory models are often evaluated by predicting what happens next from
an observed patient history. Using the same model to simulate how a visit unfolds is a
different task. During next-event prediction, each forecast is conditioned on events
that actually occurred. During simulation, generated events are fed back into the
model and become part of the history used to generate subsequent events. Errors can
therefore alter the context for later predictions and accumulate over time. This
matters in the emergency department (ED), where trajectory models could support
discharge planning, patient-flow monitoring, and staffing simulation. Yet generative EHR models are still evaluated mainly on observed
histories, next-event prediction, or downstream outcomes, which do not directly test
how a model behaves once it begins conditioning on its own outputs.

This paper examines what next-event accuracy misses. We introduce EDSim-Bench, a
closed-loop benchmark for ED trajectory simulation across two event representations
and two hospital systems. Starting from each valid prefix of a held-out visit, a model
generates the remainder of the trajectory until it produces an END event or reaches a
predefined event limit. We evaluate whether the simulated visit terminates
appropriately, whether its generated events resemble the observed continuation,
whether its timing is calibrated, whether the continuation still fits the prefix it
started from, and whether the resulting trajectories support an occupancy forecast. A train-only order-3 $n$-gram is evaluated
under the same protocol as a fixed reference.

Models that looked almost identical when predicting the next event behaved very
differently when evaluated under closed-loop generation. Across 1.13 million prefixes
from 63{,}264 held-out visits, Transformer, GRU, and LSTM next-event accuracy differed
by only 0.001, while their closed-loop event distributions and termination behaviour
varied substantially across architectures and random seeds. We also found that the
models failed in different ways: some began close to the observed next-event
distribution and drifted as generated events accumulated, while others were already
miscalibrated at the first generated step. Inference-time changes, including process
constraints, alternative sampling strategies, increased model capacity, and blending
with a count model, could improve termination without consistently restoring both the
event distribution and timing of the simulated visits. The largest improvement was
instead associated with a change in training supervision: applying the loss at every
eligible sequence position rather than only at the final position of sampled prefixes.
This pattern was observed in Transformer, GRU, and LSTM models and persisted across
model scale, temporal shift, and external replication on MC-MED. Improved stability
did not fix everything. The model with the
best event-composition score still generated visits that were too short, and model
rankings changed when the same trajectories were evaluated using occupancy
forecasting.

We make three contributions. We show that next-event accuracy can hide large
differences in rollout behaviour. We introduce a benchmark that scores termination,
event composition, timing, conditional fidelity, and occupancy under one protocol.
And we show that all-position supervision is strongly associated with more stable
rollouts across model families, while identifying what it does not fix. Clinical
trajectory models should be evaluated as simulators, across multiple seeds and
closed-loop criteria, rather than by next-event accuracy alone.

\section{Related work}
\label{sec:related}

Evaluating models under their own generated histories is a familiar problem outside
clinical data. In text generation, autoregressive models can become repetitive or
lose diversity as generation proceeds \citep{holtzman2020curious,
welleck2020unlikelihood}. In sequence learning, the mismatch between training on
observed sequences and generation from model-produced sequences is often described
as exposure bias \citep{bengio2015scheduled,ranzato2016sequence,huszar2015how}, while
work in imitation learning has shown how small one-step errors can accumulate over
longer trajectories \citep{ross2010efficient,ross2011reduction}. The same distinction
arises in learned driving systems, where low open-loop prediction error does not
guarantee good closed-loop behaviour
\citep{codevilla2018offline,dauner2023parting}. In this study, we examine both errors
that are already present at the first generated step and errors that develop
progressively as generated events are fed back into the model.

Generative EHR models such as Foresight, Delphi-2M, and ETHOS can produce
longitudinal patient trajectories, but they are still evaluated largely using
predictions made from observed histories or aggregate properties of generated patient
populations \citep{kraljevic2024foresight,shmatko2025delphi,renc2024ethos}. More
recent work has begun to compare teacher-forced evaluation with autoregressive
generation more directly \citep{mu2026ehrworld,wang2026chrono}. Temporal point
processes provide another approach by modelling event identity and timing jointly
\citep{du2016recurrent,xue2024easytpp}. We do not include them in the main
closed-loop comparison because their generation procedure was not evaluated through
the same corrected rollout framework used for the other model families in this study.

Traditional ED simulation usually represents patient flow, resources, and timing
explicitly through discrete-event or agent-based models \citep{gul2015comprehensive}.
Learned trajectory models do not, so these properties must be measured from the
trajectories they generate rather than assumed from the model structure.
EDSim-Bench therefore evaluates termination, event composition, timing,
conditional behaviour, and occupancy within a common framework and includes
replication in a second hospital system. Simple count-based sequence models also
provide useful references for this setting \citep{liu2024infinigram,nguyen2024ngram}.
We use a train-only order-3 $n$-gram as a fixed comparator throughout the benchmark
rather than treating it as an empirical optimum.

\section{The benchmark}
\label{sec:benchmark}

\subsection{Data and task}
\label{sec:data}

The internal cohort combines MIMICEL 2.1.0 and MIMIC-IV-ED 2.2
\citep{wei2023mimicel,johnson2023mimicived}, comprising 425{,}028 emergency
department stays from 205{,}466 patients and 7.99 million events after appending an
END token. We use subject-level 70/15/15 train, validation, and test splits, yielding
63{,}264 test stays and 1{,}132{,}806 valid nonterminal prefixes. For external
replication, we use MC-MED 1.0.1. After eligibility filtering and event
harmonization, 105{,}378 of 118{,}385 stays from 70{,}545 patients are retained
\citep{kansal2025mcmed}.

Each stay is represented as an ordered sequence $x_t=(a_t,\Delta t_t)$, where $a_t$
is an activity token and $\Delta t_t$ is the elapsed time since the previous event.
In open-loop evaluation, the model predicts from an observed history. In closed-loop
generation, each sampled event is added to the history and used to generate
subsequent events,
\begin{equation}
\hat{x}_{t+k}\sim p_\theta\!\left(\cdot\mid
x_{1:t},\hat{x}_{t+1:t+k-1}\right),\qquad k\geq1,
\label{eq:loop}
\end{equation}
until the model generates END. We evaluate two representations of the ED process: a
coarse 7-token vocabulary and a primary 32-token vocabulary that distinguishes order,
laboratory, medication, imaging, and disposition categories. These vocabularies
represent care processes rather than the full evolving clinical state of the patient.

\subsection{Models and training setups}
\label{sec:models}

We compare neural sequence models with several train-only process baselines. The
baselines include marginal and Markov models, order-2 and order-3 $n$-grams with
backoff, and an order-2 semi-Markov model with history-conditioned log-normal
durations. The standard neural models are a Transformer, GRU, and LSTM. Each model
encodes activity and time features and predicts the next event, time increment,
remaining length of stay (LOS), disposition, and END probability. These models are
trained using one observed prefix per example, with the loss applied at the final
position of the prefix. The causal Transformer uses the same continuous-time
prediction heads but applies causal attention and supervision at every eligible
sequence position. We also evaluate a decoder-only model in which event tokens and
quantile-binned time tokens are interleaved in a single autoregressive sequence.

\subsection{Closed-loop evaluation}
\label{sec:protocol}

For each evaluated prefix, the model generates one continuation until it produces END
or reaches a 128-event cap, corresponding to the 99.9th percentile of MIMIC visit
length. Every valid prefix is evaluated and generated events are pooled across
continuations, so longer visits and longer generations contribute more to the
estimate; sensitivity to equal-rollout weighting, starting context, and event cap is
reported in Appendices~\ref{app:lengthmatch}, \ref{app:subgroup}, and~\ref{app:cap}.
Under the vanilla protocol, audited in Appendix~\ref{app:harness}, the next event is
drawn by
categorical sampling at temperature 1 from the event softmax, in which END is an
ordinary token, and the trajectory terminates only when END is generated. For the
regression-head models the inter-event time is not sampled: it is the median of the
model's quantile head, inverted from \texttt{log1p} space, and is fed back as an
input feature; the decoder-only model instead samples binned time tokens.

\subsection{Evaluation metrics}
\label{sec:metrics}

We evaluate the following properties of the generated trajectories.
\textbf{Termination} is the proportion of continuations that generate END before
reaching the event cap, while \textbf{reached discharge} records whether a discharge
event occurs before END.

To measure event composition, we compare the pooled frequencies of generated and
observed suffix events using Jensen--Shannon divergence (JSD). This captures
differences in the overall event distribution, but not event order, timing, or
whether the continuation is appropriate for the starting context. We normalize JSD to
a train-only order-3 $n$-gram evaluated under the same protocol,
\begin{equation}
F(M)=D_{\mathrm{JSD}}(P^M_{\mathrm{gen}},P_{\mathrm{real}})/
D_{\mathrm{JSD}}(P^{n\text{-gram-3}}_{\mathrm{gen}},P_{\mathrm{real}}),
\label{eq:xf}
\end{equation}
and report this quantity as \xF{}. A value of 1\xF{} matches the fixed order-3
reference on this metric, and lower values indicate smaller divergence. Because the
denominator can be small, and because higher-order count models can improve this
score, we report raw JSD alongside \xF{} and interpret it together with the other
evaluation criteria.

Temporal calibration is summarized using the \textbf{duration ratio},
$R_{\mathrm{dur}}=\overline{\hat d}/\overline d$, where $\overline{\hat d}$ and
$\overline d$ are the mean generated and observed remaining durations, respectively.
A value of 1 indicates calibrated mean duration. We report this alongside
remaining-LOS mean absolute error because absolute error alone can favour systematic
under-generation in a right-skewed duration distribution. Additional analyses examine
edit distance (Appendix~\ref{app:closedloop}), repetition and error by rollout step
(Appendix~\ref{app:loops}), conditional and order-sensitive fidelity
(Appendices~\ref{app:ordercurve}--\ref{app:order}), a real-vs-real JSD floor
(Appendix~\ref{app:floor}), demographic variation (Appendix~\ref{app:equity}),
occupancy (Appendix~\ref{app:occupancy}), and an exploratory downstream prediction
(Appendix~\ref{app:causal}). Confidence
intervals for rates are obtained using stay-level cluster bootstrap and are reported
with the per-seed results in Appendix~\ref{app:seed}.

\begin{table}[t]
\centering
\caption{Core internal results with the 32-token vocabulary and vanilla decoding.
Panel A gives representative full-test runs; Dur. is generated over observed
remaining duration on terminated rollouts. Panel B gives seed ranges on the matched
100k-prefix audit. Raw JSD is followed by its multiple of the train-only order-3
reference. Architecture-level claims use panel B and Appendix~\ref{app:seed}.}
\label{tab:rich}
\label{tab:seed}
\footnotesize
\renewcommand{\arraystretch}{0.85}
\setlength{\tabcolsep}{5pt}
\begin{tabular}{lcccc}
\toprule
Model & Open-loop acc. & Term. & Raw JSD (\xF{}) & Dur. \\
\midrule
\multicolumn{5}{l}{\emph{A. Evaluation gap: representative full-test runs}} \\
$n$-gram-3 & -- & 0.9998 & 0.0023 (1) & 1.14 \\
Transformer & 0.442 & 0.434 & 0.310 (136) & 0.97 \\
GRU & 0.442 & 0.599 & 0.073 (32) & 0.67 \\
LSTM & 0.443 & 0.568 & 0.084 (37) & 0.45 \\
\midrule
Model (supervision) & Seeds & Term. range & Raw JSD range & \xF{} range \\
\midrule
\multicolumn{5}{l}{\emph{B. Training comparison: matched 100k-prefix seed audit}} \\
Transformer (prefix) & 7 & 0.43--0.96 & 0.010--0.314 & 4.2--137 \\
GRU (prefix) & 7 & 0.36--0.44 & 0.092--0.122 & 40--53 \\
LSTM (prefix) & 3 & 0.43--0.46 & 0.083--0.092 & 36--40 \\
Causal Transformer (prefix) & 6 & 0.65--0.95 & 0.014--0.221 & 6.2--97 \\
Decoder (all-position) & 3 & 0.9998--0.9999 & 0.0095--0.0110 & 4.1--4.8 \\
Causal Transformer (all-position) & 3 & 0.977--0.983 & 0.0015--0.0039 & 0.7--1.7 \\
GRU (all-position) & 3 & 0.982--0.985 & 0.0036--0.0136 & 1.6--6.0 \\
LSTM (all-position) & 3 & 0.981--0.985 & 0.0026--0.0063 & 1.1--2.8 \\
\bottomrule
\end{tabular}
\end{table}

\section{Open-loop and closed-loop performance}
\label{sec:gap}

\subsection{Next-event accuracy versus rollout behaviour}
\label{sec:openloop}

The three standard neural models performed almost identically when evaluated on
observed histories. With the 32-token vocabulary, next-event accuracy was 0.442 for
the Transformer and GRU and 0.443 for the LSTM. Under rollout, however, the same
models behaved very differently. In representative full-test rollouts (Appendix~\ref{app:closedloop}),
only 43.4\% of Transformer
trajectories, 59.9\% of GRU trajectories, and 56.8\% of LSTM trajectories generated
END before the event cap. Their generated event distributions also differed markedly
from the observed continuations, with raw JSD ranging from 0.073 to 0.310, compared
with 0.0023 for the train-only order-3 $n$-gram reference (Table~\ref{tab:rich},
panel A).

The difference was not limited to whether the trajectories terminated. The standard
neural models reached a discharge event in only 61--69\% of rollouts, compared with
approximately 94\% for the order-3 reference. Temporal metrics could also mislead:
with the 7-token vocabulary the neural models had lower remaining-LOS error than the
count reference while generating only 7--8\% of the observed remaining duration, and
duration ratios computed on terminated rollouts shifted substantially when
recalculated on launch points shared across models
(Table~\ref{tab:app-matchdur}).

We refer to the discrepancy between performance on observed histories and behaviour
during self-generated trajectories as the \textbf{simulation gap}. In this setting,
similar next-event accuracy did not imply similar termination, event composition, or
temporal behaviour once the models began conditioning on their own predictions.

\subsection{Variation across training seeds}
\label{sec:seedvar}

Closed-loop performance also varied considerably between independently trained
models using the same architecture and training setup. Across seven Transformer
seeds, termination ranged from 0.43 to 0.96 and raw JSD from 0.010 to 0.314,
corresponding to 4.2--137\xF{}. The recurrent models were less variable across seeds,
but consistently remained far from the count reference: GRU runs ranged from 40 to
53\xF{}, and LSTM runs from 36 to 40\xF{} (Table~\ref{tab:rich}, panel B).

A single checkpoint could therefore misrepresent a training setup: one seed was
within a few times of the count reference while another trained with the same
recipe was two orders of magnitude away, so the main closed-loop comparisons report
multiple seeds rather than a representative checkpoint.

The all-position models in Table~\ref{tab:rich} (panel B) provide a useful contrast:
termination rates were consistently close to 0.98 or higher across seeds, with much
less event-distribution divergence than the corresponding prefix-trained models.
Section~\ref{sec:regime} examines this difference under controlled comparisons.

\subsection{The count-model reference and launch context}
\label{sec:context}

The train-only order-3 $n$-gram is not a weak baseline but the reference the neural
models must reach. It terminated in 0.9998 of rollouts with mean duration ratio 1.14
(Table~\ref{tab:rich}, panel A), and every prefix-trained neural run exceeded its
event-distribution divergence by at least fourfold and most by more than thirtyfold
(panel B). Only the all-position models approached it: the causal all-position
Transformer matched it on marginal composition (0.7--1.7\xF{}) and exceeded it on
late-visit conditional fidelity (Appendix~\ref{app:ordercurve}), while the
decoder-only model, not the count model, gave the lowest occupancy error
(Appendix~\ref{app:occupancy}). Deeper count models improve marginal fidelity
further at rising storage cost (Appendices~\ref{app:ordercurve}
and~\ref{app:coverage}). Matching a trigram
on the primary criteria is therefore a minimum requirement for a simulator.

The point from which generation was started also mattered. For the 32-token
Transformer, termination was approximately 0.69--0.74 when generation began from
triage-anchored prefixes, compared with about 0.42 from other points in the visit.
These non-triage prefixes accounted for 94.5\% of the test set
(Appendix~\ref{app:subgroup}). A model evaluated
only from a single standardized starting point could therefore appear substantially
more stable than the same model evaluated throughout the course of a visit.

The gap therefore depends on the training seed and on where in the visit simulation
begins, not on any single architecture comparison. The next section examines how
the failures develop during generation.

\section{Failure modes under rollout}
\label{sec:mechanism}

The models did not fail in the same way once generation became autoregressive. With
the 7-token representation, repetition was common in the recurrent models: more than
half of GRU and LSTM rollouts contained a run of at least ten identical events. With
the 32-token representation, the Transformer showed a different pattern. Rather than
repeatedly generating a single event, it concentrated on a relatively small part of
the event vocabulary. Its unique-token ratio was 0.28, compared with 0.47 for the
order-3 $n$-gram reference, and one triage-related event increased from 0.3\% of
observed continuation events to 23.8\% of generated events.

\subsection{First-step error and compounding drift}
\label{sec:invisible}

To distinguish errors that were already present when generation began from those
that developed after the model started conditioning on its own outputs, we compared
the generated and observed event distributions at each step of the rollout
(Appendix~\ref{app:loops}). The recurrent models generally began close to the
observed next-event distribution and diverged progressively as sampled events were
added to their histories. The 32-token Transformer behaved differently: its event
distribution was already miscalibrated at the first generated step, before any
model-generated event had been fed back as input. This pattern was consistent across
training seeds, with all seven Transformer runs showing greater first-step JSD than
the three GRU runs.

Timing also changed as generation progressed, with mean inter-event intervals in
some settings contracting from approximately 6 minutes near the start of a rollout
to around 0.5 minutes by step 10, so the deterioration affected both which events
were produced and when they occurred.

The same analysis separated models that looked similar on aggregate metrics. Both
all-position models largely removed the
elevated first-step error seen in the standard Transformer, but their later behaviour
differed. The causal Transformer remained at approximately 0.002--0.004 JSD over the
first ten generated steps, whereas the decoder-only model increased to approximately
0.010--0.013 by around step 10. A low first-step error therefore does not guarantee
calibration later in the rollout.

\subsection{Inference-time interventions}
\label{sec:ceiling}

We next asked whether these closed-loop failures could be reduced without changing
how the models were trained. We tested process constraints, alternative sampling
strategies, increased model capacity, and blending neural predictions with the
order-3 $n$-gram reference. These interventions often improved termination, but
the gains did not consistently extend to event composition or timing
(Appendix~\ref{app:decode}).

Process masking is the clearest example: with the 32-token GRU, restricting
implausible transitions increased termination from 0.599 to 0.865, but the duration
ratio fell from 0.67 to 0.26; for the Transformer, masking improved termination but
had little effect on event-distribution JSD. Constraining the event sequence can make it end more often without making it
better calibrated.

Changing the sampling strategy produced similar trade-offs. In the 7-token
Transformer, a repetition penalty of 1.75 raised termination from 0.905 (vanilla) to
0.985 and reduced raw JSD from 0.091 to 0.026, but the generated trajectories were
only about 4\% as long as the observed remaining visits; in the 32-token setting,
greedy decoding was particularly unstable, with only 5.2\% of Transformer
trajectories reaching END.

We also tested whether the closed-loop failures reflected insufficient model
capacity: increasing the Transformer roughly $8\times$ in parameters improved
termination and reduced divergence, but left it close to the much smaller GRU and
well above the order-3 reference, without consistently improving temporal
calibration (Appendix~\ref{app:scale}). Finally, we blended the Transformer
predictions with those of the order-3 $n$-gram. Increasing the contribution of the
count model progressively improved termination and event-distribution JSD, but none
of the mixtures outperformed the pure count model on event composition, and the
gains in event composition were not sufficient to restore the timing of the
simulated visits.

Across these experiments, inference-time changes could correct particular
features, especially termination, but no intervention simultaneously approached
the order-3 reference on composition and timing (Figure~\ref{fig:plane}). Part of
the instability is therefore set before inference, so the next section turns to
training supervision.

\section{Training supervision and rollout stability}
\label{sec:regime}

\subsection{Prefix-only and all-position supervision}
\label{sec:attribution}

We next asked whether rollout instability was related to how the models were
trained. The standard Transformer, GRU, and LSTM were trained using
one observed prefix per example, with the loss applied only at the final position. We
compared these models with variants trained using supervision at every eligible
sequence position. The main comparisons used the same 32-token data and base model
configuration ($d=128$, two layers, feed-forward dimension 512, eight epochs), with
three to seven independent training seeds per configuration. The prefix-only models
were trained with batch size 1024 and the all-position models with batch size 256;
this difference is addressed by the batch-matched control below.

All-position training also exposes the model to more prediction targets per epoch
and raises next-event accuracy, from approximately 0.44 to 0.57 for the causal
Transformer. The controls below address architecture, causal masking, batch size,
and scheduled sampling.

\begin{figure}[t]
\centering
\includegraphics[width=\linewidth]{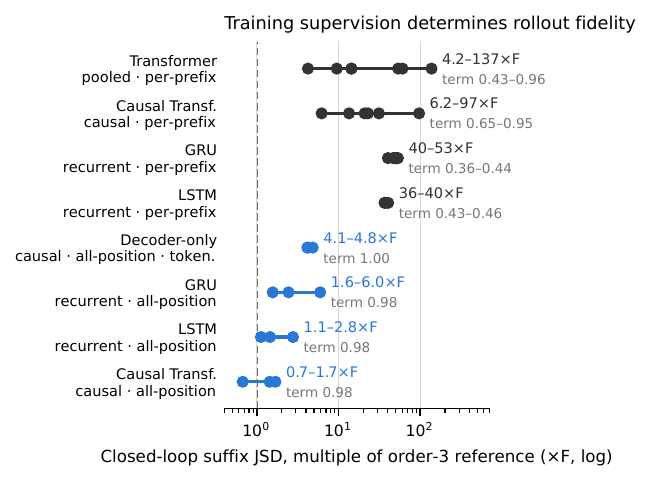}
\caption{Training supervision and closed-loop behaviour. Each point
represents one training seed using the 32-token vocabulary. Event-distribution
divergence is reported relative to the train-only order-3 $n$-gram reference
(\xF{}, log scale). Models trained with supervision at every eligible sequence
position show substantially lower divergence and less variation across seeds,
particularly for the causal Transformer.}
\label{fig:ladder}
\end{figure}

Across three seeds, the causal
all-position Transformer terminated in 0.977--0.983 of rollouts and achieved raw JSD
of 0.0015--0.0039, corresponding to 0.7--1.7\xF{}. By comparison, the standard
prefix-trained Transformer ranged from 4.2 to 137\xF{} across seven seeds, with
termination between 0.43 and 0.96. The decoder-only model, which was also trained
across all positions, terminated almost universally but remained at 4.1--4.8\xF{}.

The improvement in event composition did not extend equally to timing. The decoder
generated approximately 0.88--0.96 of the observed remaining duration, whereas the
continuous-time causal Transformer generated only about half.

\subsection{Architecture and training controls}
\label{sec:controls}

Several comparisons suggest that the improvement was associated with the supervision
setup rather than with causal attention alone. A causal Transformer trained using the
original prefix-only supervision remained highly variable, with event-distribution
divergence ranging from 6.2 to 97\xF{}, so causal masking alone was not
sufficient. The difference also survived retraining the causal all-position
Transformer at the prefix-only models' batch size of 1024: across three seeds it
achieved 1.6--3.9\xF{}.

The same pattern held in the recurrent architectures.
Changing the GRU from final-position to all-position supervision reduced divergence
from 40--53 to 1.6--6.0\xF{}, while the LSTM improved from 36--40 to 1.1--2.8\xF{}.
Termination increased from approximately 0.4 in the prefix-trained recurrent models
to approximately 0.98 in their all-position counterparts.

Scheduled sampling was not a substitute: in a 7-token Transformer it worsened both
divergence and termination relative to a matched baseline
(Appendix~\ref{app:decode}).

Across the three architecture families, all-position supervision was therefore
associated with one to two orders of magnitude lower divergence, more reliable
termination, and less seed variation, though the all-position recurrent models and
continuous-time Transformer still generated only approximately 0.32--0.53 of the
observed remaining duration.

The association held outside the primary configuration. At
approximately eight times the model size, the causal all-position Transformer
reached 1.6\xF{} with termination 0.976, compared with 26\xF{} and termination
0.751 for the same-size prefix-trained Transformer. Under temporal shift, with
models trained on 2008--2013 visits and evaluated on 2017--2019 visits, the causal
model's JSD increased by approximately eightfold, similar to the order-3 reference.
The same qualitative result replicated on MC-MED: standard Transformer and GRU
models retained high next-event accuracy but failed to terminate in any of
100{,}000 vanilla rollouts, whereas three causal all-position Transformer runs
achieved termination of 0.945--0.979 and divergence of 1.0--2.5\xF{} relative to the
site-specific order-3 reference. Scale and temporal-shift analyses are reported in
Appendix~\ref{app:scale} and the MC-MED replication in Appendix~\ref{app:external}.

\section{Discussion}
\label{sec:discussion}

Next-event prediction and trajectory simulation give very different impressions of
the same model, with large differences in termination, event composition, and seed
sensitivity once models generate complete continuations. The failures also
developed differently: some models began close to the observed next-event
distribution and drifted as their own predictions entered the history, whereas
others were already miscalibrated at the first generated step. Evaluation on
observed histories alone therefore does not establish whether a model will remain
well behaved as a simulator.

Inference-time interventions (process constraints, sampling changes, capacity,
count-model blending) could improve termination without consistently restoring
composition and timing. They are useful safeguards but not a substitute for
evaluating the full trajectory. The largest improvement was instead associated with
the training supervision scheme (Section~\ref{sec:regime}): loss at every eligible
position lowered divergence, improved termination, and reduced seed variation across
all three architectures, although rankings changed again under order-sensitive and
occupancy criteria (Appendices~\ref{app:order} and~\ref{app:occupancy}). The
vocabularies describe ED care processes rather than diagnoses or clinical state,
and MC-MED differs from MIMIC in vocabulary and visit length, so absolute
performance is compared within rather than across sites.

\section{Conclusion}

Nearly identical next-event accuracy concealed markedly different closed-loop
behaviour. Inference-time fixes improved individual failures without correcting
them together. All-position supervision was associated with substantially more
stable rollouts but did not solve every criterion. Clinical trajectory models should
therefore be evaluated through closed-loop rollout, across multiple seeds and
criteria, before predictive performance is taken as evidence of fidelity.

\label{maintextend}

\clearpage

\bibliographystyle{plainnat}
\bibliography{references}

\appendix

\section{Open-loop results}
\label{app:openloop}

Table~\ref{tab:app-openloop} reports the open-loop results underlying Section~\ref{sec:openloop}. With the primary 32-token vocabulary, the three standard architectures are nearly indistinguishable on next-event prediction: accuracy ranges from 0.442 to 0.443. With the 7-token vocabulary the spread is slightly larger, from 0.594 to 0.605, but remains small relative to the differences observed under closed-loop rollout.

The two all-position models are not directly matched to the standard models on open-loop performance. The causal all-position Transformer reaches next-event accuracy 0.567, while the decoder-only model reports 0.650. Despite this higher reported accuracy, the decoder has worse closed-loop event-distribution fidelity than the causal Transformer (mean 4.4 vs.\ 1.3\xF{}; Section~\ref{sec:attribution}). Decoder event prediction is evaluated after teacher-forcing the preceding observed time-bin token. The primary open-loop/closed-loop dissociation does not depend on this comparison and is already present among the three standard architectures.

\begin{table}[ht]
\centering
\caption{Open-loop next-event and remaining-LOS metrics. Standard-model rows use the full test split; the causal all-position and decoder-only rows use a 30k stratified sample. Decoder event accuracy is evaluated after teacher-forcing the preceding observed time-bin token and is therefore not strictly comparable with the continuous-time models.}
\label{tab:app-openloop}
\footnotesize
\setlength{\tabcolsep}{3pt}
\begin{tabular}{llccc}
\toprule
Vocab & Model & Acc. & Top-3 & LOS MAE \\
\midrule
7-tok & Transformer & 0.594 & 0.932 & 216.8 \\
7-tok & GRU & 0.604 & 0.936 & 213.3 \\
7-tok & LSTM & 0.605 & 0.937 & 213.2 \\
32-tok & Transformer & 0.442 & 0.768 & 216.6 \\
32-tok & GRU & 0.442 & 0.763 & 213.5 \\
32-tok & LSTM & 0.443 & 0.763 & 213.5 \\
32-tok & Causal Transf.\ (all-pos.) & 0.567 & 0.864 & -- \\
32-tok & Decoder-only & 0.650 & 0.896 & -- \\
\bottomrule
\end{tabular}
\end{table}

\paragraph{Temporal point processes.}
We additionally trained five neural temporal point processes (RMTPP, NHP, SAHP, THP, and AttNHP) using EasyTPP \citep{xue2024easytpp}. Their generation path is separate from the corrected rollout harness used for the other model families in this study (Section~\ref{sec:protocol}). We therefore omit their closed-loop results rather than report values that cannot be certified under the same rollout and evaluation implementation.

Temporal point processes model event identity and timing jointly, so a comparable rollout evaluation would test whether the benefit associated here with all-position sequence supervision extends to intensity-based trajectory models. That comparison requires a validated rollout policy for intensity-parameterized models rather than a direct reuse of the present discrete-event sampler, and is left to future work.

\section{Full closed-loop results}
\label{app:closedloop}

Table~\ref{tab:app-closed} reports the complete standard-model results under vanilla and process-masked decoding for both event vocabularies. The 32-token vanilla rows reproduce the single-run results summarized in Table~\ref{tab:rich}. These rows should not be interpreted as architecture-level seed averages. In particular, the GRU checkpoint shown here performs better than the seven GRU runs in the seed experiment; Table~\ref{tab:seed} and Appendix~\ref{app:seed} report the corresponding seed distributions.

Values in \xF{} are normalized to the order-3 $n$-gram count reference fitted separately within each vocabulary. Its raw JSD is 0.0016 for the 7-token vocabulary and 0.0023 for the 32-token vocabulary.

\paragraph{Duration under matched termination.}
Duration ratio is defined only for terminated rollouts, because a capped trajectory has no model-generated end time. This creates a selection problem: each model may terminate on a different subset of launch points, and a model that preferentially terminates on particular visit types can appear better calibrated for that reason.

Table~\ref{tab:app-matchdur} therefore repeats the calculation on the subset of launch points, identified by stay and prefix length, for which all four models terminate. With the 7-token vocabulary, matching changes the ratios by at most 0.01 and the severe under-generation remains. With the 32-token vocabulary, however, the standard Transformer's ratio changes from 0.97 on its own terminated rollouts to 1.18 on matched launch points. Its apparently near-calibrated aggregate ratio therefore partly reflects the subset of trajectories on which it terminates. The recurrent models move further below 1 after matching, while the count reference remains comparatively stable. We consequently do not infer the direction of temporal bias from unmatched terminated subsets alone.

\begin{table}[ht]
\centering
\caption{Duration ratio computed over each model's own terminated rollouts and over launch points where all four models terminate. The matched sets contain 288{,}146 launch points for the 32-token vocabulary and 730{,}046 for the 7-token vocabulary. A ratio of 1.00 indicates calibrated mean remaining duration.}
\label{tab:app-matchdur}
\footnotesize
\setlength{\tabcolsep}{4pt}
\begin{tabular}{llcc}
\toprule
Vocab & Model & Own term. & Matched \\
\midrule
32-tok & $n$-gram-3 & 1.14 & 1.16 \\
32-tok & Transformer & 0.97 & 1.18 \\
32-tok & GRU & 0.67 & 0.58 \\
32-tok & LSTM & 0.45 & 0.38 \\
\midrule
7-tok & $n$-gram-3 & 1.15 & 1.18 \\
7-tok & Transformer & 0.08 & 0.07 \\
7-tok & GRU & 0.07 & 0.07 \\
7-tok & LSTM & 0.07 & 0.07 \\
\bottomrule
\end{tabular}
\end{table}

\begin{table}[ht]
\centering
\caption{Full internal closed-loop performance. Van $=$ vanilla decoding; PM $=$ process-masked decoding; Dur.\ $=$ generated over observed remaining duration on terminated rollouts, with 1.00 indicating calibration; LOS $=$ prefix-stripped remaining-LOS MAE in minutes (Appendix~\ref{app:harness}); RD $=$ reached-discharge rate. LOS MAE can favor substantially miscalibrated duration, particularly in the 7-token setting. Process masking can improve termination while simultaneously worsening temporal calibration.}
\label{tab:app-closed}
\scriptsize
\setlength{\tabcolsep}{2pt}
\begin{tabular}{llcccccc}
\toprule
Vocab & Model/Dec. & Term. & JSD \xF{} & Edit & Dur. & LOS & RD \\
\midrule
7-tok & Transf.\ Van & 0.905 & 0.091 (56) & 0.679 & 0.08 & 300 & 0.623 \\
7-tok & Transf.\ PM & 0.889 & 0.088 (55) & 0.666 & 0.09 & 295 & 0.776 \\
7-tok & GRU Van & 0.680 & 0.126 (78) & 0.718 & 0.07 & 279 & 0.463 \\
7-tok & GRU PM & 0.664 & 0.125 (77) & 0.710 & 0.08 & 275 & 0.551 \\
7-tok & LSTM Van & 0.683 & 0.134 (83) & 0.718 & 0.07 & 279 & 0.466 \\
7-tok & LSTM PM & 0.668 & 0.132 (82) & 0.710 & 0.09 & 276 & 0.554 \\
7-tok & $n$-gram-3 & 0.9998 & 0.0016 (1) & 0.575 & 1.15 & 327 & 0.944 \\
\midrule
32-tok & Transf.\ Van & 0.434 & 0.310 (136) & 0.844 & 0.97 & 356 & 0.611 \\
32-tok & Transf.\ PM & 0.752 & 0.305 (133) & 0.777 & 0.64 & 295 & 0.715 \\
32-tok & GRU Van & 0.599 & 0.073 (32) & 0.811 & 0.67 & 269 & 0.694 \\
32-tok & GRU PM & 0.865 & 0.056 (25) & 0.727 & 0.26 & 267 & 0.817 \\
32-tok & LSTM Van & 0.568 & 0.084 (37) & 0.802 & 0.45 & 265 & 0.665 \\
32-tok & LSTM PM & 0.826 & 0.064 (28) & 0.728 & 0.16 & 276 & 0.782 \\
32-tok & $n$-gram-3 & 0.9998 & 0.0023 (1) & 0.656 & 1.14 & 326 & 0.943 \\
\bottomrule
\end{tabular}
\end{table}

\section{Decoding ablations}
\label{app:decode}

Tables~\ref{tab:app-dec7} and~\ref{tab:app-decR} report the complete one-knob-at-a-time decoding sweep for the Transformer and GRU. Learning-rate tuning, process masking of the tuned checkpoint, and count-model blending were evaluated for the Transformer only.

Two intended GRU evaluations with the 32-token vocabulary, greedy decoding and top-$p$ sampling, are omitted. Inspection of their rollout artifacts showed that the requested sampler configuration had not been applied correctly. We therefore exclude these cells rather than report results from an invalid intervention, consistent with the rollout-audit policy in Section~\ref{sec:protocol}.

\paragraph{Scheduled-sampling control.}
We also tested whether partial exposure to generated history repairs the per-prefix training setup. A 7-token Transformer trained with replacement probability 0.25 from the second epoch reaches 71\xF{} under process-masked decoding, compared with 55\xF{} for the matched non-scheduled baseline, while termination falls from 0.889 to 0.824. This single configuration does not rule out other schedules, but it shows that partial self-conditioning alone does not reproduce the all-position result.

\begin{table}[ht]
\centering
\caption{Decoding ablation with the 7-token vocabulary on 100k prefixes. Dur.\ is generated over observed remaining duration on terminated rollouts. No tested decoding rule brings the clock close to 1.00. For the Transformer, repetition penalty 1.75 gives the best termination and JSD of the tested settings but leaves generated duration at only 4\% of the observed remaining stay.}
\label{tab:app-dec7}
\footnotesize
\setlength{\tabcolsep}{2pt}
\begin{tabular}{lcccccc}
\toprule
& \multicolumn{3}{c}{Transformer} & \multicolumn{3}{c}{GRU} \\
\cmidrule(lr){2-4}\cmidrule(lr){5-7}
Config & Term. & JSD & Dur. & Term. & JSD & Dur. \\
\midrule
Vanilla & 0.905 & 0.091 & 0.08 & 0.680 & 0.126 & 0.07 \\
Greedy & 0.370 & 0.209 & 0.09 & 0.274 & 0.203 & 0.05 \\
Temp.\ 0.7 & 0.720 & 0.148 & 0.13 & 0.461 & 0.185 & 0.16 \\
Temp.\ 1.3 & 0.974 & 0.053 & 0.05 & 0.879 & 0.081 & 0.04 \\
Top-$p$ 0.9 & 0.461 & 0.191 & 0.16 & 0.371 & 0.199 & 0.27 \\
Rep.-pen.\ 1.2 & 0.957 & 0.064 & 0.06 & 0.825 & 0.095 & 0.04 \\
Rep.-pen.\ 1.5 & 0.981 & 0.038 & 0.05 & 0.941 & 0.059 & 0.03 \\
Rep.-pen.\ 1.75 & 0.985 & 0.026 & 0.04 & 0.972 & 0.039 & 0.03 \\
Tuned Van & 0.967 & 0.058 & 0.08 & -- & -- & -- \\
Tuned PM & 0.961 & 0.056 & 0.09 & -- & -- & -- \\
Blend 0.25 & 0.947 & 0.077 & 0.06 & -- & -- & -- \\
Blend 0.50 & 0.971 & 0.058 & 0.06 & -- & -- & -- \\
Blend 0.75 & 0.983 & 0.036 & 0.06 & -- & -- & -- \\
\bottomrule
\end{tabular}
\end{table}

\begin{table}[ht]
\centering
\caption{Decoding ablation with the 32-token vocabulary on 100k prefixes. Dur.\ is defined as in Table~\ref{tab:app-dec7}. For the GRU, each tested intervention that increases termination above the vanilla value also shortens the simulated visit. Count-model blending substantially improves Transformer termination and JSD but does not match the pure count reference.}
\label{tab:app-decR}
\footnotesize
\setlength{\tabcolsep}{2pt}
\begin{tabular}{lcccccc}
\toprule
& \multicolumn{3}{c}{Transformer} & \multicolumn{3}{c}{GRU} \\
\cmidrule(lr){2-4}\cmidrule(lr){5-7}
Config & Term. & JSD & Dur. & Term. & JSD & Dur. \\
\midrule
Vanilla & 0.434 & 0.310 & 0.97 & 0.599 & 0.073 & 0.67 \\
Greedy & 0.052 & 0.343 & 0.01 & -- & -- & -- \\
Temp.\ 0.7 & 0.271 & 0.311 & 0.83 & 0.380 & 0.125 & 0.68 \\
Temp.\ 1.3 & 0.554 & 0.300 & 0.92 & 0.709 & 0.066 & 0.48 \\
Top-$p$ 0.9 & 0.173 & 0.310 & 0.18 & -- & -- & -- \\
Rep.-pen.\ 1.2 & 0.550 & 0.318 & 0.95 & 0.703 & 0.070 & 0.44 \\
Rep.-pen.\ 1.5 & 0.650 & 0.325 & 0.84 & 0.753 & 0.084 & 0.25 \\
Rep.-pen.\ 1.75 & 0.697 & 0.328 & 0.75 & 0.775 & 0.098 & 0.19 \\
Blend 0.25 & 0.616 & 0.198 & 0.80 & -- & -- & -- \\
Blend 0.50 & 0.844 & 0.105 & 0.62 & -- & -- & -- \\
Blend 0.75 & 0.939 & 0.037 & 0.54 & -- & -- & -- \\
\bottomrule
\end{tabular}
\end{table}

\section{Scale ablation}
\label{app:scale}

Table~\ref{tab:app-scale} reports the capacity experiment discussed in Section~\ref{sec:ceiling}. With the 32-token vocabulary, increasing the standard Transformer from 0.42M to 3.25M parameters substantially improves closed-loop composition and termination: JSD falls from 0.310 to 0.059 (136 to 26\xF{}) and termination rises from 0.434 to 0.751. The larger model nevertheless remains far from the order-3 count reference and approximately at the level of the much smaller GRU on event-distribution JSD.

With the 7-token vocabulary, where the smaller Transformer already terminates in more than 90\% of rollouts, the same increase in capacity has a much smaller effect on event-distribution fidelity, from 56 to 49\xF{}. Increasing capacity also does not guarantee better temporal calibration: on matched 32-token launch points, the duration ratio moves from 1.18 for the smaller Transformer to 0.41 for the larger model (Table~\ref{tab:app-matchdur}). The corresponding all-position Transformer scale experiments are summarized in Section~\ref{sec:controls}.

\begin{table}[ht]
\centering
\caption{Scale ablation under vanilla decoding. Small (=d=128), 2 layers, 4 heads, feed-forward dimension 512 (0.42M parameters); Large (=d=256), 4 layers, 8 heads, feed-forward dimension 1024 (3.25M parameters, approximately $8\times$). The $d=512$, 6-layer all-position model summarized in Section~\ref{sec:controls} has 19.2M parameters (approximately $45\times$ the baseline). Dur.\ and LOS are defined as in Table~\ref{tab:app-closed}.}
\label{tab:app-scale}
\footnotesize
\setlength{\tabcolsep}{3pt}
\begin{tabular}{llcccc}
\toprule
Vocab & Model & Term. & JSD \xF{} & Dur. & LOS \\
\midrule
32-tok & Transf.\ small & 0.434 & 0.310 (136) & 0.97 & 356 \\
32-tok & Transf.\ large & 0.751 & 0.059 (26) & 0.41 & 271 \\
7-tok & Transf.\ small & 0.905 & 0.091 (56) & 0.08 & 300 \\
7-tok & Transf.\ large & 0.964 & 0.079 (49) & 0.15 & 301 \\
\bottomrule
\end{tabular}
\end{table}

\section{Per-seed results}
\label{app:seed}

Table~\ref{tab:app-seed} reports the individual runs underlying Table~\ref{tab:seed} and Figure~\ref{fig:ladder}. Unless stated otherwise, configurations use the same 32-token data, hyperparameters, vanilla decoding, and 100k rollout prefixes. The pooled Transformer and GRU each have seven training seeds, the pooled LSTM has three, the causal prefix-trained control has six, and the all-position configurations have three each. Seeds 1--3 are matched across the principal model comparisons.

Seeds 1--3 correspond to 20260517--20260519; Transformer and GRU seeds 4--7 correspond to 20260523--20260526; and causal prefix-control seeds 4--6 correspond to 20260520--20260522.

For normalization, \xF{} uses the 32-token full-test order-3 $n$-gram JSD of approximately 0.0023. Refitting the count reference on the matched 100k rollout sample gives approximately 0.0022 and does not materially change the reported multiples.

Min--max ranges emphasize extreme seed behavior but do not describe the distribution fully. Across runs, the standard Transformer has mean $42.0\pm47.6$\xF{} and median 14.4 ($n=7$); the standard GRU $48.9\pm4.4$\xF{} ($n=7$); and the standard LSTM $38.1\pm2.1$\xF{} ($n=3$). The all-position causal Transformer has mean $1.3\pm0.5$\xF{}, the all-position GRU $3.3\pm2.3$\xF{}, the all-position LSTM $1.8\pm0.9$\xF{}, and the decoder $4.4\pm0.4$\xF{} ($n=3$ each). The standard Transformer's large difference between its mean and median illustrates its sensitivity to random initialization and why a single checkpoint is not representative of that training setup.

Decoder rows use the common corrected stopping rule (Appendix~\ref{app:harness}). Under the earlier model-specific 24-hour guard, the same checkpoints had termination 0.952--0.959 and divergence 4.3--5.2\xF{}. Stay-level cluster-bootstrap 95\% intervals for the causal all-position Transformer and decoder do not overlap for any matched seed: [0.61, 0.74] versus [4.06, 4.34], [1.34, 1.55] versus [4.68, 4.99], and [1.60, 1.79] versus [4.02, 4.27]\xF{}.

\begin{table}[t]
\centering
\caption{Per-seed closed-loop metrics with the 32-token vocabulary and vanilla decoding. Model labels follow Table~\ref{tab:seed}: (prefix) $=$ loss at the final prefix position only; (all-pos.) $=$ supervision at every eligible position; Decoder $=$ decoder-only tokenized-time model; Cap $=$ fraction of rollouts reaching the event cap.}
\label{tab:app-seed}
\scriptsize
\renewcommand{\arraystretch}{0.88}
\setlength{\tabcolsep}{2pt}
\begin{tabular}{llccc}
\toprule
Model & Seed & Term. & JSD \xF{} & Cap \\
\midrule
Transformer & 1 & 0.432 & 0.314 (137) & 0.568 \\
Transformer & 2 & 0.901 & 0.010 (4.2) & 0.099 \\
Transformer & 3 & 0.684 & 0.138 (61) & 0.316 \\
Transformer & 4 & 0.962 & 0.033 (14) & 0.038 \\
Transformer & 5 & 0.930 & 0.033 (14) & 0.070 \\
Transformer & 6 & 0.697 & 0.122 (54) & 0.303 \\
Transformer & 7 & 0.877 & 0.022 (9.5) & 0.123 \\
\midrule
GRU & 1 & 0.395 & 0.114 (50) & 0.605 \\
GRU & 2 & 0.427 & 0.106 (46) & 0.573 \\
GRU & 3 & 0.436 & 0.112 (49) & 0.564 \\
GRU & 4 & 0.418 & 0.092 (40) & 0.582 \\
GRU & 5 & 0.364 & 0.117 (51) & 0.636 \\
GRU & 6 & 0.415 & 0.122 (53) & 0.585 \\
GRU & 7 & 0.425 & 0.119 (52) & 0.575 \\
\midrule
LSTM & 1 & 0.427 & 0.092 (40) & 0.573 \\
LSTM & 2 & 0.446 & 0.086 (37) & 0.554 \\
LSTM & 3 & 0.465 & 0.083 (36) & 0.535 \\
\midrule
Decoder (all-pos.) & 1 & 0.9999 & 0.0096 (4.2) & 0.0002 \\
Decoder (all-pos.) & 2 & 0.9998 & 0.0110 (4.8) & 0.0002 \\
Decoder (all-pos.) & 3 & 0.9998 & 0.0095 (4.1) & 0.0002 \\
\midrule
Causal Transf.\ (all-pos.) &1 & 0.981 & 0.0015 (0.7) & 0.019 \\
Causal Transf.\ (all-pos.) &2 & 0.983 & 0.0033 (1.4) & 0.017 \\
Causal Transf.\ (all-pos.) &3 & 0.977 & 0.0039 (1.7) & 0.023 \\
\midrule
GRU (all-pos.) &1 & 0.983 & 0.0036 (1.6) & 0.017 \\
GRU (all-pos.) &2 & 0.985 & 0.0056 (2.4) & 0.015 \\
GRU (all-pos.) &3 & 0.982 & 0.0136 (6.0) & 0.018 \\
\midrule
LSTM (all-pos.) &1 & 0.984 & 0.0026 (1.1) & 0.016 \\
LSTM (all-pos.) &2 & 0.981 & 0.0063 (2.8) & 0.019 \\
LSTM (all-pos.) &3 & 0.985 & 0.0033 (1.4) & 0.015 \\
\midrule
Causal Transf.\ (prefix) &1 & 0.740 & 0.221 (97) & 0.260 \\
Causal Transf.\ (prefix) &2 & 0.954 & 0.031 (13) & 0.046 \\
Causal Transf.\ (prefix) &3 & 0.655 & 0.071 (31) & 0.345 \\
Causal Transf.\ (prefix) &4 & 0.955 & 0.014 (6.2) & 0.045 \\
Causal Transf.\ (prefix) &5 & 0.930 & 0.052 (23) & 0.070 \\
Causal Transf.\ (prefix) &6 & 0.827 & 0.048 (21) & 0.173 \\
\midrule
Causal Transf.\ (all-pos., batch 1024) &1 & 0.983 & 0.0057 (2.5) & 0.017 \\
Causal Transf.\ (all-pos., batch 1024) &2 & 0.967 & 0.0036 (1.6) & 0.033 \\
Causal Transf.\ (all-pos., batch 1024) &3 & 0.986 & 0.0088 (3.9) & 0.014 \\
\bottomrule
\end{tabular}
\end{table}

\section{Higher-order count models}
\label{app:ordercurve}

The order-3 $n$-gram used throughout the paper is a fixed reference rather than an empirical optimum. To quantify how count-model performance changes with memory depth, we fit order-$k$ $n$-grams with backoff on the same training data and evaluate them on a fixed, matched sample of 100k test prefixes. Table~\ref{tab:app-order} shows a monotonic reduction in suffix event-distribution JSD through order 7, from 0.00234 at order 3 to 0.00041 at order 7. This improvement comes with rapidly increasing storage and decreasing exact-context coverage, while termination remains at or above 0.9998 for every order tested.

The result reinforces why \xF{} should be interpreted relative to a fixed order-3 reference rather than as a universal threshold for simulator validity: aggregate marginal fidelity can continue to improve simply by increasing the depth of the count table.

\begin{table}[ht]
\centering
\caption{Order-$k$ count models with the 32-token vocabulary on 100k matched prefixes. Contexts $=$ distinct training contexts stored; Size $=$ serialized table size; Cov.\ $=$ fraction of test contexts observed exactly in training at that order.}
\label{tab:app-order}
\footnotesize
\setlength{\tabcolsep}{3.5pt}
\begin{tabular}{lcccc}
\toprule
Order & JSD & Contexts & Size & Cov. \\
\midrule
3 & 0.00234 & 7{,}945 & 23\,MB & 0.9999 \\
4 & 0.00152 & 48{,}198 & 97\,MB & 0.9966 \\
5 & 0.00110 & 154{,}683 & 286\,MB & 0.9829 \\
6 & 0.00071 & 340{,}010 & 660\,MB & 0.9506 \\
7 & 0.00041 & 585{,}196 & 1.27\,GB & 0.8950 \\
\bottomrule
\end{tabular}
\end{table}

\paragraph{The same trend persists at higher event granularity.}
As an additional sensitivity analysis, we repeat the order sweep using an extended 182-token event vocabulary. The qualitative pattern is similar: order-3, order-5, and order-7 JSD are 0.00269, 0.00077, and 0.00041, respectively, with order 7 reaching approximately the same absolute marginal divergence as in the 32-token setting despite exact-context coverage falling below 58\%. The memory cost rises sharply, however: the corresponding tables occupy 0.91, 3.9, and 8.2\,GB. The order-7 table is approximately 4{,}300 times larger than the 1.9\,MB causal all-position Transformer checkpoint. Thus, deeper counting can continue to improve in-distribution marginal fidelity, but at rapidly increasing storage cost.

\paragraph{Deeper counting is more sensitive to temporal shift.}
We next refit the order-3 and order-7 models using an earlier training period (2008--2013) and evaluate them on visits from 2017--2019. Under this temporal shift, the order-7 advantage over order 3 contracts from approximately $5.7\times$ in-distribution to $1.4\times$: order-3 JSD increases from 0.0023 to 0.019, while order-7 JSD increases from 0.0004 to 0.014. Exact context coverage remains similar between the two models (90.4\% vs.\ 89.5\%), suggesting that the loss of the deeper model's advantage is not explained simply by previously unseen contexts.

The causal all-position Transformer also degrades under temporal shift, but its relative increase in JSD is approximately $8\times$, similar to the order-3 model and substantially smaller than the approximately $35\times$ increase of the order-7 model. These results are consistent with higher-order count models gaining in-distribution fidelity from increasingly specific empirical conditionals that are more sensitive to temporal changes in the underlying care process.

\paragraph{Conditional structure can favor the sequence model.}
Aggregate marginal fidelity is not the only quantity of interest. We therefore examine matched late-visit prefixes with completion fraction at least 0.75, where the continuation depends on a relatively deep observed history. Across all three causal-model seeds, the causal all-position Transformer has lower suffix-event JSD than both order-3 and order-7 count models (0.032--0.036 vs.\ 0.070 and 0.048, respectively). It also more closely matches the distribution of remaining stay, with Wasserstein-1 distance 0.09--0.12 compared with 0.63 for order 3 and 0.65 for order 7 (Table~\ref{tab:app-surv}).

These results do not imply that the neural model dominates the count models overall: deeper $n$-grams retain stronger aggregate event-frequency fidelity, and the semi-Markov model remains strong on explicit duration modeling. Rather, the late-visit analysis suggests that longer-context sequence representations can preserve aspects of state-conditional trajectory structure that are not captured as well by the finite-order count models tested here.

\begin{table}[ht]
\centering
\caption{Conditional fidelity on matched late-visit prefixes (completion fraction $\geq0.75$). JSD $=$ suffix event-distribution JSD within this subset; LOS-W1 $=$ Wasserstein-1 distance in log-minutes between generated and observed remaining-stay \emph{distributions}. LOS-W1 is a distributional rather than paired per-visit metric and is therefore compatible with the global duration biases reported in Table~\ref{tab:rich}. Causal-model values span three training seeds.}
\label{tab:app-surv}
\footnotesize
\setlength{\tabcolsep}{3.5pt}
\begin{tabular}{lcc}
\toprule
Model & JSD & LOS-W1 \\
\midrule
$n$-gram-3 & 0.070 & 0.63 \\
$n$-gram-7 & 0.048 & 0.65 \\
Semi-Markov & -- & 0.37 \\
Causal all-position & 0.032--0.036 & 0.09--0.12 \\
\bottomrule
\end{tabular}
\end{table}

\paragraph{Conditional two-sample stress test.}
The headline JSD compares marginal suffix-event distributions and does not test whether a generated continuation is appropriate for the specific prefix from which it was launched. We therefore construct a conditional classifier two-sample test (C2ST). For each launch point, we form a pair consisting of the observed prefix and either its real or generated continuation. A gradient-boosted classifier is trained under grouped 3-fold cross-validation first using continuation features alone (\emph{marginal} AUC) and then using features from both the prefix and continuation (\emph{conditional} AUC).

Because real and generated examples are paired on the same launch prefixes, a prefix-only classifier has AUC 0.500. The increase from marginal to conditional AUC therefore measures the additional separability available when the continuation is evaluated together with the state from which it was generated (Table~\ref{tab:app-c2st}). It should not be interpreted as a complete or model-independent measure of conditional distributional error.

The order-3 and order-7 count models have nearly identical conditional increments despite their substantially different marginal JSDs. Their raw C2ST gaps are both approximately $+0.036$, suggesting that deeper counting improves aggregate composition more than it improves this measure of prefix-conditioned fidelity.

Raw conditional-minus-marginal gaps are difficult to compare directly because models with larger marginal AUC have less remaining headroom before AUC 1. We therefore additionally normalize the gap by available headroom,
\begin{equation}
\frac{\mathrm{AUC}_{\mathrm{cond}}-\mathrm{AUC}_{\mathrm{marg}}}
{1-\mathrm{AUC}_{\mathrm{marg}}},
\label{eq:headroom}
\end{equation}
and report an analogous logit-scale increment. Under these corrections, the causal all-position model has a somewhat smaller conditional increment than the order-3 count model: 0.40--0.44 versus 0.49 by headroom normalization and 0.53--0.59 versus 0.71 on the logit scale. The standard pooled Transformer instead has the largest corrected conditional increment in the table.

An important tension remains. The causal all-position model has marginal C2ST AUC 0.967--0.977, compared with 0.927 for the count models, meaning that the classifier distinguishes its generated suffixes from real suffixes more easily using the engineered continuation features even though the causal model matches or exceeds the count reference on token-histogram JSD. The metrics evaluate different properties: JSD uses marginal event frequencies, whereas the C2ST can exploit a broader feature representation. We therefore do not claim that the causal model's generated continuations are indistinguishable from real data.

\begin{table}[ht]
\centering
\caption{Conditional classifier two-sample test on 100k matched prefixes using a gradient-boosted classifier with grouped 3-fold cross-validation. Marg.\ and Cond.\ are classifier AUC using continuation-only and prefix-plus-continuation features, respectively. Head.\ is the conditional increment normalized by available AUC headroom; Logit is the corresponding increment on the logit scale. Lower corrected increments indicate less additional separability after conditioning on the prefix. Causal-model values span three training seeds.}
\label{tab:app-c2st}
\scriptsize
\setlength{\tabcolsep}{3pt}
\begin{tabular}{lcccc}
\toprule
Model & Marg. & Cond. & Head. & Logit \\
\midrule
$n$-gram-3 & 0.927 & 0.962 & 0.49 & 0.71 \\
$n$-gram-7 & 0.927 & 0.963 & 0.49 & 0.71 \\
Semi-Markov & 0.950 & 0.976 & 0.52 & 0.77 \\
Causal all-position & 0.967--0.977 & 0.980--0.987 & 0.40--0.44 & 0.53--0.59 \\
\midrule
Standard Transformer & 0.998 & 0.999 & 0.78 & 1.50 \\
\bottomrule
\end{tabular}
\end{table}

\section{Sensitivity to event order}
\label{app:order}

The primary suffix event-distribution JSD operates on token histograms. It therefore measures whether the generated suffix contains events in approximately the correct proportions, but is insensitive to their ordering. A model could in principle reproduce the correct marginal event frequencies while arranging those events unrealistically.

We test this limitation by recomputing suffix-distribution divergence over unigrams, bigrams, and trigrams. For each order, model divergence is normalized to the order-3 count model evaluated using the same $n$-gram statistic, so that 1\xF{} again denotes the corresponding count reference at that order.

\begin{table}[ht]
\centering
\caption{Suffix-distribution divergence under increasingly order-sensitive representations. Values are multiples of the order-3 count reference scored at the same sequence order. The standard models use the full test set; all-position and decoder rows use the matched 100k-prefix evaluation.}
\label{tab:app-order-sens}
\scriptsize
\setlength{\tabcolsep}{3pt}
\begin{tabular}{lccc}
\toprule
Arm & $n=1$ & $n=2$ & $n=3$ \\
\midrule
$n$-gram-3 reference & 1.0 & 1.0 & 1.0 \\
Transformer & 136 & 119 & 79 \\
GRU & 32 & 46 & 37 \\
LSTM & 37 & 52 & 40 \\
\midrule
$n$-gram-3 reference & 1.0 & 1.0 & 1.0 \\
Causal Tf.\ (all-position) & 0.7 & 1.1 & \textbf{1.4} \\
GRU (all-position) & 1.5 & 3.4 & 3.5 \\
LSTM (all-position) & 1.1 & 6.0 & \textbf{6.2} \\
Decoder (all-position) & 4.1 & 9.5 & 8.8 \\
\bottomrule
\end{tabular}
\end{table}

Two conclusions follow. First, the negative results are robust to event order: the collapsing standard models remain one to two orders of magnitude above the count reference under unigram, bigram, and trigram scoring. Their closed-loop failure is therefore not an artifact of using an order-insensitive headline metric.

Second, near-parity on marginal event composition does not necessarily imply near-parity in local sequential structure. The causal all-position Transformer remains close to the count reference, moving from 0.7\xF{} on unigrams to 1.4\xF{} on trigrams. Other all-position models separate more substantially: the LSTM moves from 1.1 to 6.2\xF{}, the GRU from 1.5 to 3.5\xF{}, and the decoder from 4.1 to 8.8\xF{}. Thus, marginal and order-sensitive fidelity constitute distinct evaluation axes and can rank otherwise similar models differently. We retain unigram JSD as the primary composition metric for consistency with the fixed count reference, but release the bigram- and trigram-based variants alongside it.

\section{Sensitivity to rollout weighting}
\label{app:lengthmatch}

The primary suffix JSD pools event tokens over all generated suffixes and compares their aggregate histogram with the corresponding pooled histogram from the observed suffixes. This weighting gives longer generated trajectories more influence. A capped rollout can contribute up to 128 generated events, whereas an observed remaining visit is typically much shorter. Consequently, models with high non-termination may contribute disproportionate mass from long capped trajectories.

Across 35 seed-level runs, cap fraction and \xF{} are strongly associated (Spearman $\rho=+0.84$). We therefore test whether the primary composition result is simply a consequence of non-termination by rescoring the same rollouts in three ways: (i) the primary pooled-token metric; (ii) a per-rollout metric in which each continuation contributes equally regardless of length; and (iii) a diagnostic restricted to the first ten events on each side.

\begin{table}[ht]
\centering
\caption{Sensitivity of event-distribution divergence to rollout weighting. Values are multiples of the order-3 count reference scored using the same weighting and the same rollout sample. ``Per-rollout'' gives each continuation equal weight regardless of length. The first-10 metric is included only as a sensitivity analysis because its count-reference denominator is extremely small.}
\label{tab:app-lengthmatch}
\scriptsize
\setlength{\tabcolsep}{3pt}
\begin{tabular}{lccc}
\toprule
Arm & Pooled & Per-rollout & First 10 \\
\midrule
Transformer (prefix) & 136 & 361 & 4445 \\
GRU (prefix) & 32 & 99 & 449 \\
LSTM (prefix) & 36 & 83 & 291 \\
\midrule
Causal Tf.\ (all-position) & \textbf{0.6} & 2.3 & 15 \\
LSTM (all-position) & 1.0 & \textbf{2.2} & 15 \\
GRU (all-position) & 1.4 & 3.3 & 37 \\
Decoder (all-position) & 4.1 & 28 & 131 \\
\bottomrule
\end{tabular}
\end{table}

For each column, \xF{} is normalized by the order-3 count model evaluated under the same weighting on the same sample. These denominators therefore differ slightly from the full-test count reference used for the main results. For example, the same causal-Transformer seed is 0.6\xF{} in the pooled analysis here and approximately 0.7\xF{} in the matched occupancy analysis (Table~\ref{tab:app-occupancy}); the difference reflects the corresponding sample-specific count denominator and rounding. The main-text value of 1.3\xF{} is different again because it reports the mean across three training seeds rather than this single seed.

Equalizing rollout weights does not make the collapsing neural models appear better. Instead, their divergence increases: the standard Transformer, for example, moves from 136 to 361\xF{}. The large separation between the prefix-trained and all-position models therefore cannot be explained solely by long non-terminating trajectories dominating the pooled histogram. The qualitative result is preserved, although some close model comparisons change under equal weighting.

This sensitivity analysis does qualify claims of ``parity'' with the count reference. The causal all-position Transformer is below 1\xF{} under pooled-token weighting but at 2.3\xF{} when each rollout receives equal weight. Matching the order-3 reference should therefore be understood as a statement about a specified metric and weighting convention rather than a property of the model in general. We retain pooled-token JSD as the primary composition metric for consistency across the benchmark and report the alternative weighting alongside it. We do not interpret the first-10 multiples quantitatively because the corresponding count-reference JSD is approximately $6\times10^{-5}$, making the normalized ratio highly sensitive to small absolute differences.

\section{Train-context coverage}
\label{app:coverage}

Table~\ref{tab:app-cov} reports the fraction of test contexts observed in the training data at increasing $n$-gram order. Through order 3, exact coverage is nearly complete for both event vocabularies, although the number of distinct contexts rises rapidly with event granularity and history length.

Coverage alone does not determine count-model fidelity. As shown in Appendix~\ref{app:ordercurve}, increasing $n$-gram order beyond 3 continues to improve marginal event-distribution JSD even as exact-context coverage falls, because the backoff model can revert to shorter observed histories when a higher-order context is unavailable.

\begin{table}[ht]
\centering
\caption{Exact train-context coverage of test transitions. Contexts are the distinct histories stored in the corresponding training count table.}
\label{tab:app-cov}
\footnotesize
\begin{tabular}{llcc}
\toprule
Vocab & Order & Coverage & \# contexts \\
\midrule
7-tok & 1 & 1.000000 & 7 \\
7-tok & 2 & 1.000000 & 26 \\
7-tok & 3 & 0.999996 & 86 \\
32-tok & 1 & 1.000000 & 31 \\
32-tok & 2 & 0.999998 & 576 \\
32-tok & 3 & 0.999853 & 7{,}945 \\
\bottomrule
\end{tabular}
\end{table}

\section{Rollout-harness audit}
\label{app:harness}

Closed-loop evaluation depends on the implementation of the rollout itself as well as on the trained model. We therefore audited the generation and evaluation pipeline across all model families before freezing the results reported in this manuscript. The audit identified six implementation inconsistencies affecting earlier experimental runs. All six were corrected and covered by regression tests before the analyses reported here.

Three issues affected the generation path. First, an earlier Transformer rollout configuration applied an undocumented heuristic decoding stack consisting of top-$k=4$, repetition penalty 1.75, a self-loop cap of 3, and an END-head threshold of 0.35 rather than the vanilla decoding protocol subsequently adopted for the benchmark. Second, an earlier rollout implementation did not correctly advance the conditioning history as generation proceeded. Third, generated timing values were not transformed back from the model's \texttt{log1p} representation using \texttt{expm1}, producing artificially short inter-event intervals. These issues affected earlier estimates of termination and timing and motivated the common decoding and generation protocol used throughout the final benchmark.

The subsequent cross-model audit identified three additional metric inconsistencies. Because rollout tables are produced by several generation paths, we explicitly checked whether each metric had the same definition regardless of which producer generated the trajectory and whether it measured only model-generated behavior after the launch prefix.

\paragraph{Remaining-LOS boundary.}
The remaining-LOS calculation originally summed time deltas over the entire stored trajectory rather than only over the generated suffix. This affected model families differently because the process-baseline producer zero-padded prefix deltas whereas the neural-model producers retained the observed historical deltas. The resulting quantity therefore represented remaining duration for the process models but elapsed-plus-remaining duration for the neural models. After stripping the observed prefix consistently, Transformer remaining-LOS MAE changes from 621 to 356 minutes and its generated-versus-observed rank correlation from (-0.16) to (+0.16); for the causal all-position Transformer, MAE changes from 335 to 239 minutes and rank correlation from (-0.05) to (+0.40). All remaining-LOS results in this manuscript use the corrected suffix-only definition.

\paragraph{Decoder stopping rule.}
The decoder-only rollout path previously applied an additional 24-hour simulated-time stopping rule that was not used by the other model families. This rule was reached in approximately 4.5\% of decoder rollouts, compared with 0.008\% reaching the shared 128-event cap. Under the common stopping rule, decoder termination increases from approximately 0.954 to 0.9998. All decoder comparisons reported here use the shared stopping criterion.

\paragraph{Reached-discharge boundary.}
The reached-discharge metric originally searched the complete stored trajectory, including the observed prefix, for a discharge token. Launch points at which discharge had already appeared in the prefix could therefore be credited as model-generated discharges. This was especially visible on MC-MED: all four standard-model cells returned reached-discharge 0.011 because 1.08\% of launch prefixes already contained a discharge event, including configurations with model-driven termination exactly 0.000. Restricting the metric to the generated suffix moves these MC-MED values to 0.000 and reduces the corresponding internal values by approximately 0.04--0.12. All reached-discharge results in the manuscript use the suffix-only definition.

The composition analyses were unaffected by these three metric-boundary corrections because suffix JSD, edit distance, and rollout-diversity statistics already remove the observed prefix before scoring. Separately, two decoding-ablation cells were excluded after artifact validation showed that their requested sampler settings had not been applied correctly (Appendix~\ref{app:decode}).

All corrected paths are regression-tested in \texttt{tests/}, including explicit tests for remaining-LOS and reached-discharge boundary handling. We additionally report generated-to-observed duration ratio alongside absolute timing error throughout the benchmark because the latter can favor systematic under-generation of right-skewed visit durations.

Table~\ref{tab:app-harness} provides one illustrative before-and-after comparison for the 7-token Transformer. Because multiple generation changes separate the earlier and corrected runs, these values summarize the cumulative audit rather than attributing the difference to any single defect.

\begin{table}[ht]
\centering
\caption{Illustrative earlier versus corrected rollout results for the 7-token Transformer. `Earlier'' uses the pre-audit generation path; `Corrected'' uses the common rollout protocol reported in this manuscript.}
\label{tab:app-harness}
\footnotesize
\begin{tabular}{lcc}
\toprule
Metric & Earlier & Corrected \\
\midrule
Termination & 0.564 & 0.905 \\
Non-termination & 0.436 & 0.095 \\
Inter-event timing & implementation artifact & artifact removed \\
\bottomrule
\end{tabular}
\end{table}

The audit illustrates a broader methodological point: closed-loop benchmarks require the same implementation discipline as model training. When several model families use different generation paths, a metric can appear numerically plausible while representing different quantities across models. Shared stopping rules, explicit prefix boundaries, cross-producer metric tests, and regression-tested rollout code are therefore part of reproducible simulator evaluation rather than incidental implementation details.

\section{Downstream length-of-stay prediction}
\label{app:causal}

As an exploratory analysis, we asked whether information available from the observed ED trajectory could support downstream flow prediction without using future events. We predict top-decile total length of stay (LOS) at fixed completion-fraction landmarks using only delays observed before each landmark. Performance ranges from AUROC 0.72--0.78 internally and 0.70--0.78 on MC-MED. These trajectory-derived features therefore contain reproducible signal across sites, but do not outperform a baseline using elapsed time alone at any landmark.

We additionally evaluate discharge-readiness prediction. Discrimination reaches AUROC 0.804 internally but falls to 0.564 on MC-MED. Local recalibration improves calibration but not discrimination. We report these analyses as exploratory examples of what process-level trajectory features can and cannot support, rather than as deployment results or evidence of clinical utility.

\section{Occupancy forecasting}
\label{app:occupancy}

The benchmark metrics measure intrinsic properties of generated trajectories. We therefore test whether their rankings carry over to a simple operational query: among patients currently in the ED, what fraction remain present after $h$ hours?

For each rollout, a patient is considered present at horizon $h$ when the generated remaining duration exceeds $h$. If a rollout reaches the event cap without terminating, its discharge time is unknown; for this analysis we conservatively count that patient as remaining present at all evaluated horizons. This convention makes non-termination directly costly rather than silently truncating the simulated visit. Occupancy is the fraction of launch points predicted to remain present, and error is reported in percentage points relative to observed occupancy. Positive values indicate overprediction of census. Models are compared only on shared launch points under the same rollout protocol.

\begin{table}[ht]
\centering
\caption{Occupancy forecast error in percentage points (predicted minus observed) at five horizons and mean absolute error across horizons. Top: full test set with 1.13M matched launch points. Bottom: matched 100k-prefix comparison of the count reference and two all-position models.}
\label{tab:app-occupancy}
\scriptsize
\setlength{\tabcolsep}{2pt}
\begin{tabular}{lccccccc}
\toprule
Arm & \xF & +1h & +2h & +4h & +8h & +12h & MAE \\
\midrule
$n$-gram-3 & 1 & (+1.4) & (+2.9) & (+6.6) & (+8.9) & (+4.5) & 4.9 \\
Transformer & 136 & (+2.2) & (+10.1) & (+26.7) & (+48.5) & (+53.2) & 28.1 \\
GRU & 32 & (-7.9) & (-2.1) & (+12.3) & (+30.0) & (+33.2) & 17.1 \\
LSTM & 37 & (-7.4) & (-5.9) & (+6.2) & (+27.1) & (+34.1) & 16.1 \\
\midrule
$n$-gram-3 & 1 & (+1.4) & (+2.9) & (+6.9) & (+9.2) & (+4.7) & 5.0 \\
Causal all-position & 0.7 & (-5.8) & (-12.4) & (-18.9) & (-13.3) & (-7.9) & 11.7 \\
Decoder & 4.2 & (-1.5) & (-2.3) & (-2.8) & (-1.4) & (-1.8) & \textbf{2.0} \\
\bottomrule
\end{tabular}
\end{table}

The operational ranking differs from the event-composition ranking. Across the seven evaluated arms, non-termination is strongly associated with occupancy error (Spearman $\rho=+0.93$), whereas the association with \xF{} is weaker ($\rho=+0.63$). These correlations are computed across the seven model configurations.

Among the four configurations with termination above 0.95, temporal calibration tracks occupancy more closely than event composition. The decoder has $|R_{\mathrm{dur}}-1|=0.09$ and occupancy MAE 2.0 percentage points; the order-3 count reference has corresponding values 0.14 and 4.9; and the causal all-position Transformer 0.47 and 11.7.

Most importantly, the composition ranking reverses for this particular operational task. The causal all-position Transformer has the best event-composition score at 0.7\xF{} but the largest occupancy error among the reliably terminating models, whereas the decoder scores 4.2\xF{} on composition but has the lowest occupancy error. Choosing a model by event-distribution JSD alone would therefore select the poorer occupancy forecaster in this experiment.

Occupancy is one downstream query, and non-terminating rollouts are handled using an explicitly conservative convention. The analysis illustrates that intrinsic composition fidelity and task-specific utility are distinct evaluation axes.

\section{Sensitivity to launch context}
\label{app:subgroup}

Closed-loop performance varies with the point in the observed trajectory from which simulation begins. Table~\ref{tab:app-subgroup} reports 7-token termination according to whether the launch prefix ends at the triage event and, when it does, the recorded triage acuity.

Acuity is available for this analysis only when the launch prefix terminates exactly at triage. The acuity-specific columns therefore contain one triage-anchored launch point per eligible stay: 62{,}220 prefixes from the 98.4\% of test stays with recorded triage acuity. The ``Non-triage'' group contains every other launch point, totaling 1{,}070{,}586 of 1{,}132{,}806 test prefixes (94.5\%). It should therefore be interpreted as a launch-context contrast rather than a patient subgroup or a pure early-versus-late comparison.

\begin{table}[ht]
\centering
\caption{Termination rate with the 7-token vocabulary by launch context under vanilla decoding. A1--A5 correspond to prefixes ending at the triage event with the indicated recorded acuity. ``Non-triage'' contains all other launch points ($n=1{,}070{,}586$; 94.5\% of test prefixes).}
\label{tab:app-subgroup}
\footnotesize
\setlength{\tabcolsep}{1.5pt}
\begin{tabular}{lcccccc}
\toprule
Model & A1 & A2 & A3 & A4 & A5 & Non-triage \\
\midrule
Transformer & 0.994 & 0.996 & 0.996 & 0.996 & 1.000 & 0.899 \\
GRU & 0.536 & 0.528 & 0.523 & 0.523 & 0.515 & 0.689 \\
LSTM & 0.457 & 0.444 & 0.441 & 0.447 & 0.412 & 0.697 \\
\bottomrule
\end{tabular}
\end{table}

The direction of the launch-context effect depends on architecture. The Transformer terminates in at least 0.994 of triage-anchored rollouts but only 0.899 of non-triage-anchored rollouts. The recurrent models show the opposite pattern, with termination around 0.41--0.54 from triage-anchored prefixes and approximately 0.69 from other launch points. A simulator evaluated only from a single standardized launch point could therefore give a substantially different impression of closed-loop stability from one evaluated throughout the visit.

Termination also declines across observed trace-length quartiles for the 7-token Transformer, from 0.957 in the shortest quartile to 0.857 in the longest. This pattern is consistent with greater difficulty on longer trajectories, although trace length also captures differences in visit complexity and case mix and therefore does not isolate compounding error. With the 32-token vocabulary, the Transformer similarly terminates more often from triage-anchored prefixes than from other launch points (approximately 0.71 vs.\ 0.42).

The released evaluation outputs additionally report cluster-bootstrap subgroup estimates across launch-context variables including triage acuity, arrival transport, chief complaint, trace-length quartile, and completion fraction.

\section{Performance by recorded demographic group}
\label{app:equity}

We additionally examine whether closed-loop performance differs across recorded demographic groups. This analysis is descriptive. Differences in model performance between groups may reflect case mix, documentation, visit stage, care processes, sampling, or model behavior, and should not be interpreted as evidence of a causal demographic effect or algorithmic discrimination.

Table~\ref{tab:app-equity} reports termination, reached-discharge rate, and duration ratio for the 32-token standard models stratified by recorded race. The duration ratio is calculated separately within each group as mean generated remaining duration divided by mean observed remaining duration. This normalization accounts for group differences in mean observed duration, but does not eliminate other forms of case-mix or trajectory-distribution differences between groups.

\begin{table}[ht]
\centering
\caption{Closed-loop performance with the 32-token vocabulary and vanilla decoding by recorded race. $n$ denotes rollout prefixes; Term.\ $=$ termination rate, RD $=$ reached-discharge rate, and Dur.\ $=$ within-group duration ratio. ``Not recorded'' is a documentation category and is shown separately rather than interpreted as a demographic group.}
\label{tab:app-equity}
\scriptsize
\setlength{\tabcolsep}{3pt}
\begin{tabular}{lrccc}
\toprule
Recorded group & $n$ & Term. & RD & Dur. \\
\midrule
\multicolumn{5}{l}{\emph{Transformer}} \\
Asian & 34{,}611 & 0.474 & 0.666 & 1.130 \\
Black & 240{,}372 & 0.431 & 0.599 & 0.927 \\
Hispanic/Latino & 87{,}113 & 0.437 & 0.609 & 1.024 \\
White & 693{,}761 & 0.432 & 0.607 & 0.944 \\
Not recorded & 16{,}097 & 0.437 & 0.709 & 2.039 \\
\midrule
\multicolumn{5}{l}{\emph{GRU}} \\
Asian & 34{,}611 & 0.625 & 0.688 & 0.708 \\
Black & 240{,}372 & 0.582 & 0.685 & 0.630 \\
Hispanic/Latino & 87{,}113 & 0.592 & 0.688 & 0.694 \\
White & 693{,}761 & 0.600 & 0.698 & 0.666 \\
Not recorded & 16{,}097 & 0.718 & 0.750 & 1.455 \\
\midrule
\multicolumn{5}{l}{\emph{LSTM}} \\
Asian & 34{,}611 & 0.613 & 0.660 & 0.467 \\
Black & 240{,}372 & 0.555 & 0.657 & 0.424 \\
Hispanic/Latino & 87{,}113 & 0.566 & 0.660 & 0.458 \\
White & 693{,}761 & 0.565 & 0.667 & 0.447 \\
Not recorded & 16{,}097 & 0.700 & 0.721 & 0.904 \\
\bottomrule
\end{tabular}
\end{table}

Among the recorded race categories, the same descriptive ordering appears across the three standard architectures: the Black group has the lowest duration ratio and reached-discharge rate, while the Asian group has the highest duration ratio. The magnitude varies by model. For the Transformer, duration ratios range from 0.927 in the Black group to 1.130 in the Asian group; differences are smaller for the recurrent models.

This consistency is worth reporting but is not sufficient to establish a systematic fairness disparity. Multiple rollout prefixes originate from the same stays, demographic groups differ in clinical and trajectory characteristics, and the present table does not adjust for those differences. A stronger fairness analysis would require stay-level uncertainty estimates and comparisons conditioned on clinically relevant visit characteristics and launch context.

Differences by recorded sex are substantially smaller in these experiments, with termination varying by less than 0.025 and duration ratio by less than 0.018 across the compared groups.

The ``Not recorded'' category behaves differently from all recorded groups, particularly for duration ratio, and likely captures a distinct mixture of documentation and visit characteristics. We therefore do not interpret it demographically. We also do not report a demographic comparison for the count reference because the corresponding rollout artifacts do not contain the required demographic linkage. Extending the same stratified evaluation to count-based and all-position models would be an important follow-up.

\section{Real-versus-real JSD reference}
\label{app:floor}

To determine whether suffix marginal JSD is inherently large because real ED futures are stochastic, we estimate the value obtained when real suffixes appear on both sides of the comparison. In each of five trials, we independently sample two sets of 100k observed test suffixes, matching the scale of the neural-model evaluations, pool their suffix-event histograms, and apply the same \texttt{js\_divergence} calculation used throughout the benchmark.

Table~\ref{tab:app-floor} reports the mean real-vs-real JSD together with population-versus-subsample variability, the order-3 count reference, and the standard vanilla Transformer. Two independent real samples differ by only $4.5\times10^{-6}$ with the 7-token vocabulary and $1.0\times10^{-5}$ with the 32-token vocabulary. The metric therefore admits values near zero at the sample sizes used here. The vanilla Transformer's divergence is more than four orders of magnitude larger in both settings, so its large JSD cannot be explained by ordinary real-sample variability alone. This comparison is a calibration reference for the metric rather than a claim that every valid stochastic simulator should reproduce the exact observed marginal.

\begin{table}[ht]
\centering
\caption{Reference values for suffix event-distribution JSD. Real vs.\ real compares two independent 100k samples of observed test suffixes and reports the mean across five trials. ``Neural'' is the standard vanilla Transformer.}
\label{tab:app-floor}
\footnotesize
\begin{tabular}{lcc}
\toprule
Quantity & 7-token & 32-token \\
\midrule
Real vs.\ real & $4.5\times10^{-6}$ & $1.0\times10^{-5}$ \\
Population vs.\ subsample & $1.5\times10^{-6}$ & $4.2\times10^{-6}$ \\
$n$-gram-3 reference & 0.0016 & 0.0023 \\
Transformer (vanilla) & 0.091 & 0.310 \\
\bottomrule
\end{tabular}
\end{table}

\section{Sensitivity to the rollout cap}
\label{app:cap}

The primary MIMIC rollouts use a 128-event generation cap, corresponding to approximately the 99.9th percentile of observed visit length. Because termination rate necessarily depends on the available generation budget, we repeat the standard vanilla rollouts with a 256-event cap.

Doubling the cap increases termination for the 7-token GRU from 0.680 to 0.815 and for the LSTM from 0.683 to 0.801 (Table~\ref{tab:app-cap}). A substantial fraction of their cap-128 failures therefore represent trajectories that continue generating for more than 128 events rather than trajectories that would necessarily repeat indefinitely. The repetition diagnostics in Appendix~\ref{app:loops} separately quantify persistent low-diversity and repeated-token behavior near the end of generated trajectories.

The principal closed-loop dissociation remains under the larger cap. The recurrent models still terminate substantially less often than the order-3 count reference, while the 32-token Transformer's termination changes only from 0.434 to 0.447. Termination should therefore be interpreted as a cap-dependent metric and compared only under a common disclosed stopping rule.

The appropriate cap is also dataset-specific. MC-MED visits are substantially longer than MIMIC visits, with approximately 34\% exceeding 128 events. We therefore use a 512-event cap for MC-MED, chosen using the same approximately 99.9th-percentile rule rather than carrying the MIMIC threshold across datasets.

\begin{table}[ht]
\centering
\caption{Termination under 128- and 256-event caps with vanilla decoding.}
\label{tab:app-cap}
\footnotesize
\begin{tabular}{llcc}
\toprule
Vocab & Model & Cap 128 & Cap 256 \\
\midrule
7-token & Transformer & 0.905 & 0.979 \\
7-token & GRU & 0.680 & 0.815 \\
7-token & LSTM & 0.683 & 0.801 \\
32-token & Transformer & 0.434 & 0.447 \\
\bottomrule
\end{tabular}
\end{table}

\section{Repetition and per-step error}
\label{app:loops}

\begin{figure}[t]
\centering
\includegraphics[width=\linewidth]{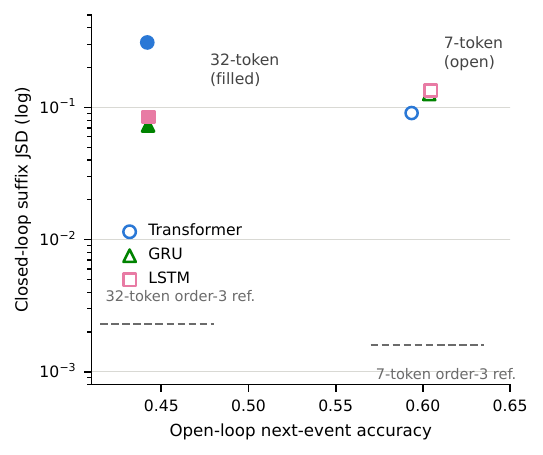}
\caption{The simulation gap. Open-loop next-event accuracy ($x$-axis) versus closed-loop suffix event-distribution JSD (log $y$-axis) for the three standard architectures. With the 32-token vocabulary, nearly identical open-loop accuracy corresponds to substantially different closed-loop divergence. Open markers denote the 7-token vocabulary and filled markers the 32-token vocabulary; the order-3 count reference is dashed.}
\label{fig:gap}
\end{figure}

\begin{figure}[t]
\centering
\includegraphics[width=0.92\textwidth]{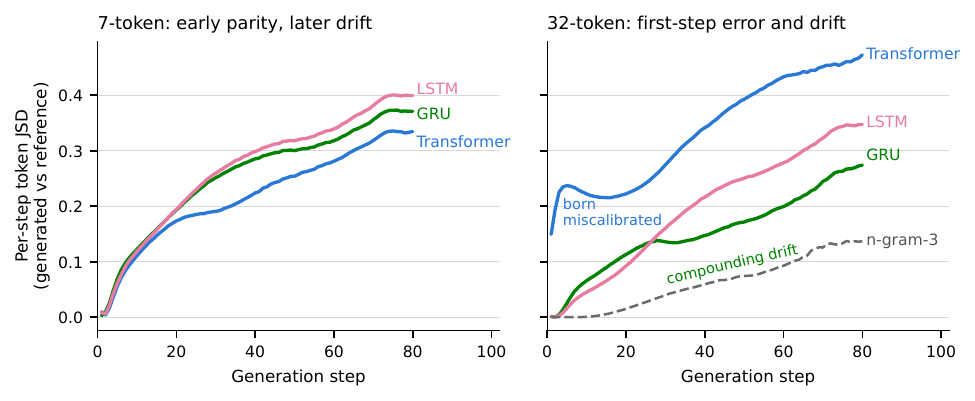}
\caption{Per-step generated-versus-observed token JSD. Left, 7-token vocabulary: the standard models begin near the observed distribution and diverge with rollout depth. Right, 32-token vocabulary: the Transformer shows elevated \emph{first-step miscalibration}, whereas the recurrent models begin closer to the observed distribution and exhibit \emph{compounding drift}. The order-3 $n$-gram reference (dashed) also accumulates some divergence with rollout depth but remains lowest.}
\label{fig:perstep}
\end{figure}

Figure~\ref{fig:gap} visualizes the simulation gap between open-loop next-event accuracy and closed-loop event-distribution fidelity. Table~\ref{tab:loop} quantifies repeated and low-diversity generation, while Figure~\ref{fig:perstep} examines how event-distribution error develops with rollout depth.

\begin{figure}[t]
\centering
\includegraphics[width=0.92\textwidth]{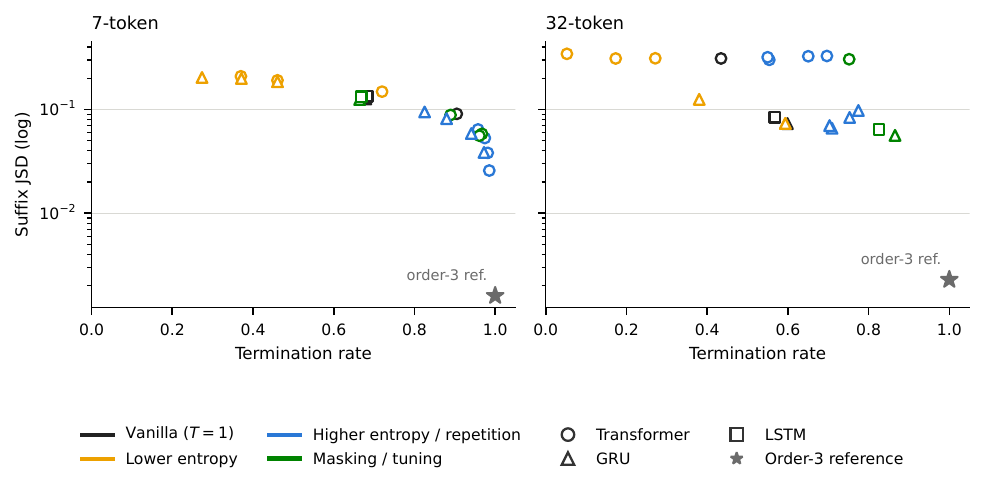}
\caption{The grammar--statistics split (Section~\ref{sec:ceiling}). Decoding, masking, and tuning interventions are shown in the termination--suffix-JSD plane for both event vocabularies. Colors identify intervention families and marker shapes identify model architectures. Several interventions improve termination and, particularly with the 7-token vocabulary, reduce event-distribution divergence, but none simultaneously reaches near-complete termination and the order-3 reference's composition. Temporal calibration is evaluated separately in Table~\ref{tab:app-closed}.}
\label{fig:plane}
\end{figure}

\begin{table}[t]
\centering
\caption{Repeated- and low-diversity-generation diagnostics under vanilla decoding on the full test split. Run $=$ mean longest identical-token run per rollout; $\geq10$ $=$ fraction containing an identical-token run of at least ten events; Uniq $=$ number of distinct generated tokens divided by suffix length; Tail $=$ fraction whose final ten generated events are identical.}
\label{tab:loop}
\footnotesize
\setlength{\tabcolsep}{4pt}
\begin{tabular}{llcccc}
\toprule
Vocab & Model & Run & $\geq10$ & Uniq & Tail \\
\midrule
7-token & Transformer & 9.0 & 0.40 & 0.43 & 0.011 \\
7-token & GRU & 11.6 & 0.57 & 0.34 & 0.038 \\
7-token & LSTM & 12.5 & 0.57 & 0.34 & 0.060 \\
32-token & Transformer & 4.4 & 0.05 & 0.28 & 0.007 \\
32-token & GRU & 6.9 & 0.24 & 0.30 & 0.025 \\
32-token & LSTM & 5.3 & 0.09 & 0.32 & 0.003 \\
32-token & $n$-gram-3 & 4.2 & 0.04 & 0.47 & 0.001 \\
\bottomrule
\end{tabular}
\end{table}

\paragraph{First-step miscalibration is seed-robust but varies in magnitude.}
The elevated first-step error of the standard 32-token Transformer is not specific to the representative checkpoint in Table~\ref{tab:rich}. Under the matched 100k-prefix evaluation, all seven standard Transformer seeds have larger first-step JSD than all three GRU seeds. Transformer values are 0.016, 0.032, 0.034, 0.046, 0.048, 0.072, and 0.156, compared with 0.003, 0.005, and 0.006 for the GRU.

The separation is therefore robust across the tested seeds, although its magnitude varies considerably: the Transformer median is 0.046, while the worst seed reaches 0.156. We consequently interpret the phenomenon as \emph{first-step miscalibration} rather than attaching the conclusion to the magnitude of any single run. The recurrent models consistently begin much closer to the observed first-step distribution and accumulate larger errors only later in the rollout.

The same analysis distinguishes the two all-position models. The causal all-position Transformer starts at approximately 0.0015 JSD across seeds and the tokenized-time decoder at approximately 0.0003, both substantially below the standard Transformer seeds. Their later behavior diverges. By approximately step 10, the causal model remains around 0.002--0.004, whereas the decoder rises to approximately 0.010--0.013. The decoder's remaining closed-loop gap is therefore consistent with error accumulating under self-conditioning rather than an elevated error already present at the first generated event.

This analysis motivates using the first generated step together with the full per-step curve as a diagnostic: it distinguishes error already visible before self-conditioning from error that develops progressively during rollout.

\section{External-site replication}
\label{app:external}

We repeat the training comparison on MC-MED using its harmonized event vocabulary. The standard Transformer and GRU and the causal all-position Transformer are each retrained from scratch on MC-MED using their respective internal training setups. All models are then evaluated on 100k prefixes using the site-specific 512-event cap.

Under vanilla decoding, both standard models have termination exactly 0.000 and reached-discharge rate 0.000 (Table~\ref{tab:mcmed}). Process masking produces an apparent termination rate of 0.011, but these 1{,}080 rollouts correspond to launch prefixes in which discharge had already occurred; under the transition mask, END is then the only legal next event. No other masked rollout terminates. Process masking therefore produces no model-driven recovery of termination.

Because the standard models do not terminate, their suffix JSD is calculated over capped trajectories and is not length-matched to the terminating count and all-position models. These JSD values should consequently be interpreted together with termination rather than as direct standalone comparisons.

The external collapse is not simply explained by a failure to learn next-event structure from MC-MED. On observed prefixes, the same standard checkpoints achieve next-event accuracy 0.689 for the Transformer and 0.719 for the GRU, substantially above their corresponding MIMIC values. High open-loop performance therefore coexists with complete failure to terminate when these checkpoints are rolled out under their own predictions.

In contrast, the three causal all-position Transformer runs terminate in 0.945--0.979 of rollouts and reach discharge in 0.927--0.947, with event-distribution divergence of 1.0--2.5\xF{} relative to an order-3 count reference refitted on MC-MED. This reproduces the qualitative training-setup association on an external hospital system, while differences in dataset size, vocabulary, and visit-length distribution preclude direct comparison of absolute performance between MIMIC and MC-MED.

\begin{table}[ht]
\centering
\caption{MC-MED replication on 100k prefixes with a 512-event cap. \xF{} is normalized to the train-only MC-MED order-3 $n$-gram reference (JSD 0.0041649), not to the MIMIC denominator. Standard-model JSD is computed over capped, non-terminating rollouts and is therefore not length-matched to the terminating rows. RD $=$ reached-discharge rate. Only the Transformer and GRU standard architectures were retrained on MC-MED.}
\label{tab:mcmed}
\scriptsize
\setlength{\tabcolsep}{3pt}
\begin{tabular}{llccc}
\toprule
Model & Decoding & Term. & JSD \xF{} & RD \\
\midrule
$n$-gram-3 reference & -- & 0.991 & 0.00416 (1) & -- \\
\midrule
Transformer & Vanilla & 0.000 & 0.0236 (5.7) & 0.000 \\
Transformer & Process mask & 0.011 & 0.0206 (4.9) & 0.000 \\
GRU & Vanilla & 0.000 & 0.0713 (17) & 0.000 \\
GRU & Process mask & 0.011 & 0.0649 (16) & 0.000 \\
\midrule
Causal all-pos.\ s1 & Vanilla & 0.945 & 0.0103 (2.5) & 0.934 \\
Causal all-pos.\ s2 & Vanilla & 0.979 & 0.0042 (1.0) & 0.947 \\
Causal all-pos.\ s3 & Vanilla & 0.963 & 0.0089 (2.1) & 0.927 \\
\bottomrule
\end{tabular}
\end{table}

\end{document}